\documentclass[10pt,twocolumn,letterpaper]{article}

\usepackage{cvpr}              

\usepackage{graphicx}
\usepackage{amsmath}
\usepackage{amssymb}
\usepackage{booktabs}
\usepackage[accsupp]{axessibility} 

\usepackage[pagebackref,breaklinks,colorlinks]{hyperref}

\usepackage[capitalize]{cleveref}
\crefname{section}{Sec.}{Secs.}
\Crefname{section}{Section}{Sections}
\Crefname{table}{Table}{Tables}
\crefname{table}{Tab.}{Tabs.}

\def\cvprPaperID{2} 
\def\confName{CVPR}
\def\confYear{2023}

\begin{document}

\title{1st Place Solution for MOSE Track in CVPR 2024 PVUW Workshop: Complex Video Object Segmentation}

\author{Deshui Miao\textsuperscript{\rm 1, \rm 2 \footnotemark[3]}, Xin Li\textsuperscript{\rm 2}, Zhenyu He\textsuperscript{\rm 1, \rm 2}, Yaowei Wang\textsuperscript{\rm 2} and Ming-Hsuan Yang\textsuperscript{\rm 3}\\
\textsuperscript{\rm 1}Harbin Institute of Technology, Shenzhen \quad 
\textsuperscript{\rm 2} Peng Cheng Laboratory\\ \quad 
\textsuperscript{\rm 3} University of California at Merced\\ 
Team: PCL\_VisionLab \\}

\maketitle
\renewcommand{\thefootnote}{\fnsymbol{footnote}}
\footnotetext[3]{Work done during research internships at Peng Cheng Laboratory.}

\begin{abstract}
Tracking and segmenting multiple objects in complex scenes has always been a challenge in the field of video object segmentation, especially in scenarios where objects are occluded and split into parts. In such cases, the definition of objects becomes very ambiguous. The motivation behind the MOSE dataset is how to clearly recognize and distinguish objects in complex scenes. In this challenge, we propose a semantic embedding video object segmentation model and use the salient features of objects as query representations. The semantic understanding helps the model to recognize parts of the objects and the salient feature captures the more discriminative features of the objects. Trained on a large-scale video object segmentation dataset, our model achieves first place (\textbf{84.45\%}) in the test set of PVUW Challenge 2024: Complex Video Object Segmentation Track.
\end{abstract}

\begin{figure}[t]
\centering
\includegraphics[width=1\columnwidth]{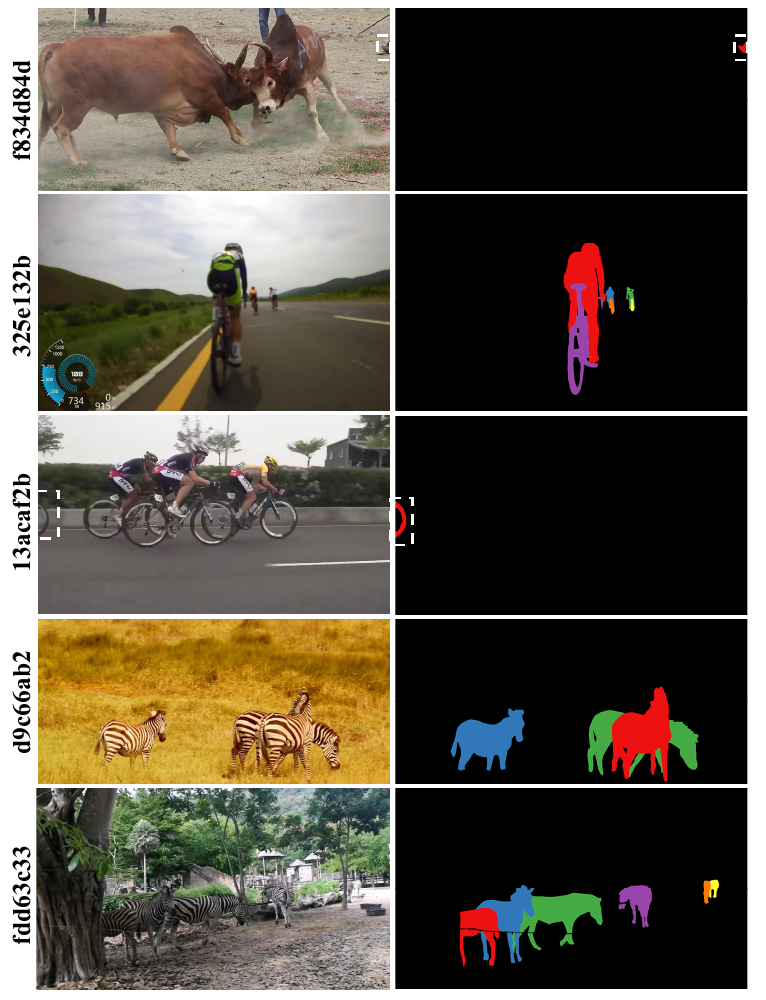}
\caption{Visual examples of the MOSE test set~\cite{mose}. The left is the first frame and the right is the corresponding targets.}
\label{fig:dataset}
\end{figure}

\begin{figure*}[t]
\centering
\includegraphics[width=1\textwidth]{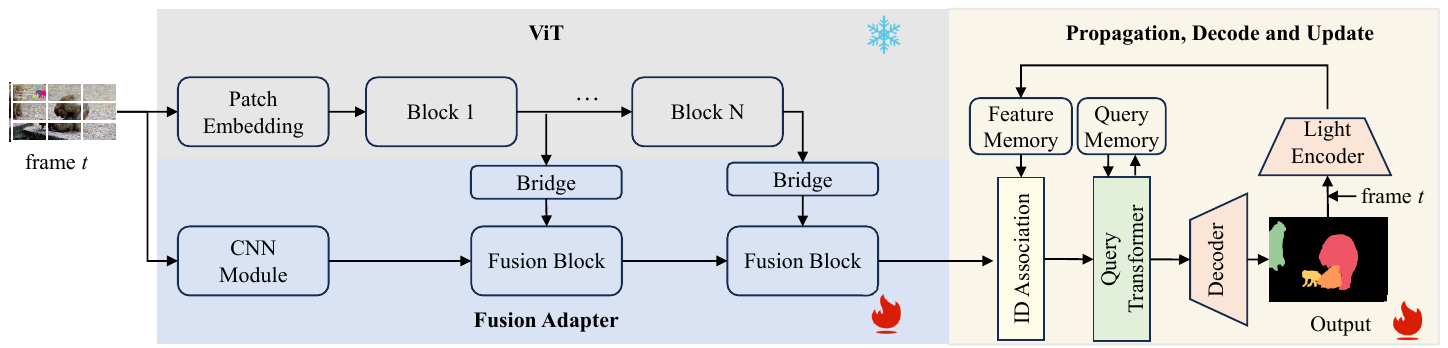}
\caption{Overall framework of our methods.}
\label{fig:overall}
\end{figure*} 

\section{Introduction}
Video Object Segmentation (VOS) focuses on tracking and segmenting target objects within a video sequence, beginning with mask annotations in the initial frame~\cite{ytvos2018, davis2017, feelvos, aot}. 
This technique has significant potential across various applications, especially given the increasing volume of video content in areas such as autonomous driving, augmented reality, and interactive video editing~\cite{perclip, bao2018cnnmrf}.
The current VOS task faces several major challenges, including significant changes in target appearance, occlusions, and identity confusion due to similar objects and background clutter. These challenges are particularly pronounced in long-term videos, making the task even more complex.

VOS methods~\cite{STM, stcn} generally achieve video object segmentation by comparing the test frame with previous frames. 
They begin by employing an association model to generate pixel-wise correlated features between the test frame and target templates. 
These correlated features are then used to predict the target masks accurately~\cite{SimVOS}.
To handle variations in target appearances over time, certain methods~\cite{aot, xmem} incorporates a memory module that stores samples of the targets as their appearances change.
Additionally, recent approaches~\cite{ISVOS, cutie} introduce object queries to differentiate between various target objects, thus mitigating identity confusion.

To evaluate the performance of current methods in realistic environments, The MOSE~\cite{mose} (coMplex video Object SEgmentation) dataset is a comprehensive collection designed to provide complex and challenging scenarios with occluded and crowded objects. 
Unlike existing datasets that feature relatively isolated and salient objects, MOSE focuses on realistic environments, highlighting the limitations of current VOS methods and encouraging the development of more robust algorithms. 
This dataset is crucial for advancing VOS technologies to perform effectively in real-world applications where object interactions and occlusions are common.

Current methods struggle with the MOSE dataset due to several challenges. First, when targets have multiple complex or separate parts because of occlusion, background clutter, and shape complexity, existing methods tend to produce incomplete prediction masks. This issue arises because these methods rely heavily on pixel-level correlation, which emphasizes pixel-wise similarity while overlooking semantic information. Second, although object queries~\cite{ISVOS, cutie} improve ID association accuracy, they perform poorly in sequences with significant target appearance changes and very small targets. The current query propagation strategy, which updates using the entire online predicted target sample, is susceptible to introducing noise and accumulating errors. This can lead to tracking failures, such as missing targets and ID switches, particularly when tracking small targets.

To address the aforementioned issues, we propose a VOS framework to learn both the semantic prior and discriminative query representation.
Specifically, we design a block that efficiently learns both semantic and detailed information, which can extract rich semantic features from a pre-trained Vision Transformer (ViT) without the need to train all feature extraction parameters.
To better model the query representation of the target, we design a more discriminative filtering mechanism to generate discriminative queries.

The main designs in this project include: 
\begin{itemize}
\item we propose a fusion block to incorporate the semantic information from the ViT feature for Video Object Segmentation. We use the cls token from a pretrained ViT backbone to extract the semantic prior and global average polling of the image patches to extract the global information of the current frames. Furthermore, we design a local fusion to leverage the spatial information from the ViT features. The proposed framework shows robustness when facing multiple complex sequences.
\item We develop a discriminative feature filtering module to model the discriminative query representation which achieves great improvement when dealing with small objects.
\end{itemize}

\begin{figure*}[t]
\centering
\includegraphics[width=1\textwidth]{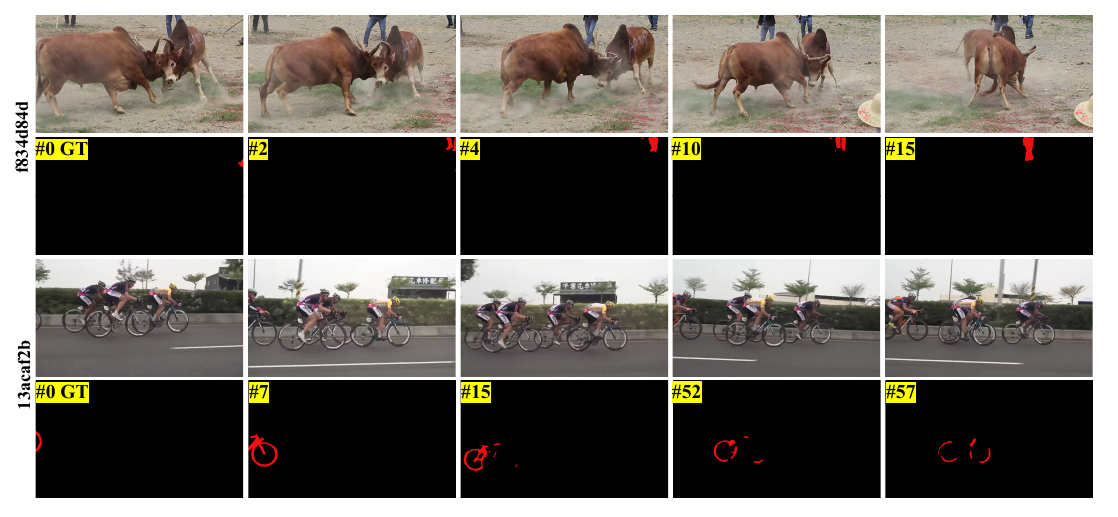}
\vspace{-0.5cm}
\caption{Performace on sequences with small targets.}
\label{fig:small}
\end{figure*} 

\section{Our solution}
To solve the problems of VOS, we propose a robust semantic-aware and query-enhanced video object segmentation method.
In this solution, we first introduce the proposed fusion block, which utilizes the semantic and detailed information of the pretrained ViT models. This helps us deal with complex target appearance variance and ID confusion between targets with similar appearances. In detail, we fuse the information of the cls token from the ViT to multi-scale features and conduct local fusion between frame patches and multi-scale features for detailed fusion. In addition, to ensure the target representation of the target queries, we develop a discriminative query representation module in the query transformer to capture the local representation of the targets. Details of the proposed solution are described as follows.

\subsection{Fusion Block}
Since the VOS task involves generic objects without class labels, learning semantic representations directly from the VOS dataset during training is challenging. 
However, the CLS token in a pre-trained ViT captures semantic information from the entire image, providing a comprehensive, global representation of the image content. 
By integrating the CLS token with multi-scale features generated from CNN networks, we can acquire detailed semantic features at various scales.
In Figure~\ref{fig:overall}, cross-attention is used to perform semantic prior learning for VOS.

Then, multi-scale deformable cross-attention is utilized to learn the spatial dependence of different scale features, which helps handle objects with complex structures or separate parts.

\subsection{Discriminative Query Generation}
We note that updating the target query memory directly with entire object patches generated based on online predicted masks is ineffective as the predicted masks often cover background noise, reducing target distinctiveness and leading to accumulating errors over time.
To propagate target queries effectively across frames, we update the target queries with the most distinctive feature of the target object.

In detail, we select the discriminative feature of a target object by comparing the target query with every channel activation in the correlated feature map of the target and taking the most similar one.
Based on the discriminative target feature generated from a new target sample, we can update target queries by dynamically calculating the relationship between the salient query and salient pixel features in an additive manner.
The proposed discriminative query generation scheme adaptively refines target queries with the most representative features, which helps deal with the challenges of dramatic appearance variations in long-term videos.
\begin{table}
\centering
\caption{Ranking results (Top 6) in the MOSE test set. We mark our results in {\color{blue}{blue}}.}
\setlength{\tabcolsep}{2mm}{
\renewcommand\arraystretch{1.0}
\begin{tabular}{c | c | c c c}
\toprule
Rank & Name & $\mathcal{J}\&\mathcal{F}$ & $\mathcal{J}$ & $\mathcal{F}$ \\
\midrule
1 & \textcolor{blue}{PCL\_MDS} &\textcolor{blue}{84.45} &\textcolor{blue}{81.01} &\textcolor{blue}{87.89} \\
2 & Yao\_Xu\_MTLab &83.45 &80.07 &86.83\\
3 & ISS &82.19 &78.79 &85.59 \\
4 & xsong2023 &82.09 &78.73 &85.44 \\
5 & yangdonghan50 &81.39 &77.99 &84.80 \\
6 & YongxinWang &80.64 &77.24 &84.04 \\

\bottomrule
\end{tabular}
}
\label{tab:performance}
\end{table}

\begin{figure*}[t]
\centering
\includegraphics[width=1\textwidth]{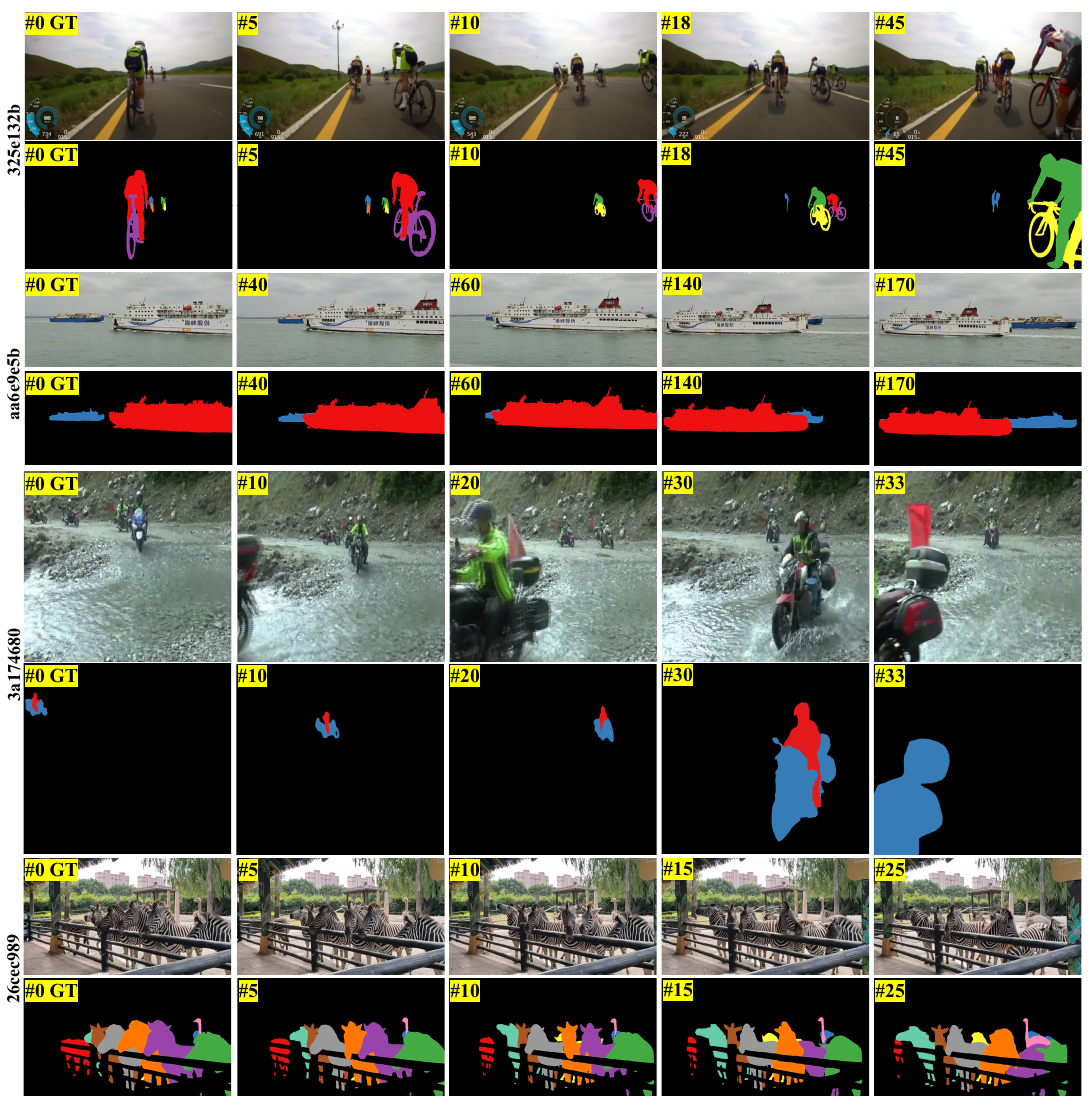}
\vspace{-0.5cm}
\caption{Qualitative results on complex sequences.}
\label{fig:result}
\end{figure*} 
\section{Experiments}
\subsection{Challenge Description}
The Pixel-level Video Understanding in the Wild (PVUW) Challenge features four tracks. This year, two new tracks: the Complex Video Object Segmentation Track, based on MOSE~\cite{mose}, and the Motion Expression Guided Video Segmentation Track, based on MeViS~\cite{mevis} are included. These new tracks include additional videos and annotations that present challenging elements such as object disappearance and reappearance, small inconspicuous objects, heavy occlusions, and crowded environments in MOSE. Furthermore, the MeViS dataset is provided to explore natural language-guided video understanding in complex settings. These enhancements aim to promote the development of more comprehensive and robust pixel-level video scene understanding in complex and realistic scenarios.

\subsection{Implementation Details}

\noindent \textbf{Training}. 
Our training settings are similar to Cutie~\cite{cutie}. 
To enhance the performance of our model, we utilize the MEGA dataset constructed by Cutie, which includes the YouTubeVOS~\cite{ytvos2018}, DAVIS~\cite{davis2017}, OVIS~\cite{ovis}, MOSE~\cite{mose}, and BURST~\cite{burst} datasets.
We sample eight frames to train the model, and three are randomly selected to train the matching process.
For each sequence, we randomly choose at most three targets for training.
The point supervision in loss is adopted to reduce the memory requirements.
We train the model for 195k on the MEGA dataset.
All our models are trained on 8 x NVIDIA V100 GPUs and tested on an NVIDIA V100 GPU.

\noindent\textbf{Inference.} 
Our feature and query memory is updated every 3rd frame during the testing phase. 
For longer sequences, we employ a long-term fusion strategy~\cite{xmem} for updating. To enhance storage quality, we skip frames without targets and do not store them.
The test input size contains two scales: 720 for general size and 1080 for small targets.
The final score is a version of multi-scale fusion~\cite{xmem}.

\noindent \textbf{Evaluation Metrics.} 
We use mean Jaccard $\mathcal{J}$ index and mean boundary
$\mathcal{F}$ score, along with mean $\mathcal{J}\&\mathcal{F}$ to evaluate segmentation accuracy. 

\subsection{Results}
The proposed solution achieves 1st place on the complex video object segmentation track of the PVUW Challenge 2024, as listed in~\ref{tab:performance}. In addition, we also show some of our quantitative results in Fig.~\ref{fig:result} and Fig.~\ref{fig:small}. It can be seen that the proposed solution can accurately segment small targets and distinguish similar targets in some difficult scenarios which have severe changes in object appearance, and confusion of multiple similar objects and small objects.
In the five submissions, we find that some inference parameters influence the performance, which are the test size, the memory interval, memory or not, the flip augmentation, and multi-scale fusion.









\section{Conclusion}
\label{sec:Conclusion}
In this paper, we propose a robust solution for the task of video object segmentation, which helps the model understand the semantic information of the targets and generate discriminative queries of the target. 
In the end, we achieve 1st place on the complex video object segmentation track of the PVUW Challenge 2024 with 84.45\% $\mathcal{J}\&\mathcal{F}$.
The detailed version is under peer review.
The code and full version will be released as soon as possible.
{\small
\bibliographystyle{plain}
\bibliography{egbib}

\begin{thebibliography}{10}

\bibitem{burst}
Ali Athar, Jonathon Luiten, Paul Voigtlaender, Tarasha Khurana, Achal Dave, Bastian Leibe, and Deva Ramanan.
\newblock Burst: A benchmark for unifying object recognition, segmentation and tracking in video.
\newblock In {\em WACV}, pages 1674--1683, 2023.

\bibitem{bao2018cnnmrf}
L.~Bao, B.~Wu, and W.~Liu.
\newblock Cnn in mrf: Video object segmentation via inference in a cnn-based higher-order spatiotemporal mrf.
\newblock In {\em CVPR}, 2018.

\bibitem{xmem}
H.~K. Cheng and A.~G. Schwing.
\newblock Xmem: Long-term video object segmentation with an atkinson-shiffrin memory model.
\newblock In {\em ECCV}, 2022.

\bibitem{stcn}
H.~K. Cheng, Y.~W. Tai, and C.~K. Tang.
\newblock Rethinking spacetime networks with improved memory coverage for efficient video object segmentation.
\newblock In {\em NeurIPS}, pages 11781--11794, 2021.

\bibitem{cutie}
Ho~Kei Cheng, Seoung~Wug Oh, Brian Price, Joon-Young Lee, and Alexander Schwing.
\newblock Putting the object back into video object segmentation.
\newblock {\em arXiv preprint arXiv:2310.12982}, 2023.

\bibitem{mevis}
Henghui Ding, Chang Liu, Shuting He, Xudong Jiang, and Chen~Change Loy.
\newblock Mevis: A large-scale benchmark for video segmentation with motion expressions.
\newblock In {\em CVPR}, pages 2694--2703, 2023.

\bibitem{mose}
Henghui Ding, Chang Liu, Shuting He, Xudong Jiang, Philip~HS Torr, and Song Bai.
\newblock Mose: A new dataset for video object segmentation in complex scenes.
\newblock In {\em ICCV}, pages 20224--20234, 2023.

\bibitem{STM}
Seoung~Wug Oh, Joon-Young Lee, Ning Xu, and Seon~Joo Kim.
\newblock Video object segmentation using space-time memory networks.
\newblock In {\em ICCV}, pages 9226--9235, 2019.

\bibitem{perclip}
Kwanyong Park, Sanghyun Woo, Seoung~Wug Oh, In~So Kweon, and Joon-Young Lee.
\newblock Per-clip video object segmentation.
\newblock In {\em CVPR}, 2022.

\bibitem{davis2017}
J.~Pont-Tuset, F.~Perazzi, S.~Caelles, P.~Arbeláez, A.~Sorkine-Hornung, and L.~Van~Gool.
\newblock The 2017 davis challenge on video object segmentation.
\newblock {\em arXiv preprint arXiv:1704.00675}, 2017.

\bibitem{ovis}
Jiyang Qi, Yan Gao, Yao Hu, Xinggang Wang, Xiaoyu Liu, Xiang Bai, Serge Belongie, Alan Yuille, Philip~HS Torr, and Song Bai.
\newblock Occluded video instance segmentation: A benchmark.
\newblock {\em IJCV}, 130(8):2022--2039, 2022.

\bibitem{feelvos}
P.~Voigtlaender, Y.~Chai, F.~Schroff, H.~Adam, and L.C. Chen.
\newblock Feelvos: Fast end-to-end embedding learning for video object segmentation.
\newblock In {\em CVPR}, pages 9481--9490, 2019.

\bibitem{ISVOS}
Junke Wang, Dongdong Chen, Zuxuan Wu, Chong Luo, Chuanxin Tang, Xiyang Dai, Yucheng Zhao, Yujia Xie, Lu~Yuan, and Yu-Gang Jiang.
\newblock Look before you match: Instance understanding matters in video object segmentation.
\newblock In {\em CVPR}, 2023.

\bibitem{SimVOS}
Qiangqiang Wu, Tianyu Yang, Wei Wu, and Antoni Chan.
\newblock Scalable video object segmentation with simplified framework.
\newblock In {\em ICCV}, 2023.

\bibitem{ytvos2018}
Ning Xu, Linjie Yang, Yuchen Fan, Dingcheng Yue, Yuchen Liang, Jianchao Yang, and Thomas Huang.
\newblock Youtube-vos: A large-scale video object segmentation benchmark.
\newblock In {\em ECCV}, 2018.

\bibitem{aot}
Zongxin Yang, Yunchao Wei, and Yi~Yang.
\newblock Associating objects with transformers for video object segmentation.
\newblock In {\em NeurIPS}, 2021.

\end{thebibliography}
}

\end{document}